\documentclass[fleqn,10pt,lineno]{wlpeerj}
\usepackage{makecell}
\usepackage{longtable}
\usepackage{multirow}
\title{Suicidal Ideation Detection on Social Media: A Review of Machine Learning Methods}

\author[1]{Asma Abdulsalam}
\author[1]{Areej Alhothali}
\affil[1]{Department of Computer Science,Faculty of Computing and Information Technology,King AbdulAziz University,Jeddah,Saudi Arabia}

\corrauthor[1]{Asma Abdulsalam}{aabdulsalam0012@stu.kau.edu.sa}

\begin{abstract}
Social media platforms have transformed traditional communication methods by allowing users worldwide to communicate instantly, openly, and frequently. People use social media to express their opinion and share their personal stories and struggles. Negative feelings that express hardship, thoughts of death, and self-harm are widespread in social media, especially among young generations. Therefore, using social media to detect and identify suicidal ideation will help provide proper intervention that will eventually dissuade others from self-harming and committing suicide and prevent the spread of suicidal ideations on social media. Many studies have been carried out to identify suicidal ideation and behaviors in social media. This paper presents a comprehensive summary of current research efforts to detect suicidal ideation using machine learning algorithms on social media. This review 24 studies investigating the feasibility of social media usage for suicidal ideation detection is intended to facilitate further research in the field and will be a beneficial resource for researchers engaged in suicidal text classification.
\end{abstract}

\begin{document}

\flushbottom
\maketitle
\thispagestyle{empty}

\section*{Introduction}

Millions of individuals regularly use social media such as chat rooms, blogging websites, and social networking platforms, with $3.96$ billion people actively utilizing the internet~\cite{1astoveza2018suicidal}. Facebook, Twitter, Snapchat, and other social media networking sites allow users to share material and interact with others. Many users prefer to utilize social media networks to share their thoughts and emotions, and their daily experiences, problems, and issues. Suicidal ideation, death, and self-harming thoughts are among the most widely discussed themes on social media.

Suicide is described as a person's deliberate attempt to take their own life~\cite{2nock2008suicide}. Suicide is a multifaceted occurrence that results from a complex interaction of biological, psychological, social, cultural, and spiritual variables~\cite{3beck1979assessment}. Suicide is a manifestation of underlying suffering caused by a mix of events, including underlying mental diseases that generate psychological pain \cite{4liu2020suicidal}. Suicide ideation, suicide planning, and suicide attempts are three types of suicidal behavior~\cite{2nock2008suicide,3beck1979assessment,4liu2020suicidal}. Suicide ideation refers to a person's ideas or intentions to end their life without actually trying to do so. In contrast, a suicide plan is a specific technique a person can use to end their life, and a suicide attempt is an act of self-harm that results in death with the intended purpose being to die~\cite{2nock2008suicide,3beck1979assessment,4liu2020suicidal}.

Suicide has ramifications for people, families, communities, and even countries~\cite{4liu2020suicidal}. Suicide is the second largest cause of mortality among young people, killing more people than diabetes, liver disease, stroke, or infection~\cite{5weber2017psychiatric}. More than 40\% of individuals who seek primary care are reluctant to address their depressive symptoms because of the stigma associated with mental disorders. Suicidal thoughts and acts necessitate quick intervention, and there is no reliable approach for managing, assessing, or preventing suicide~\cite{5weber2017psychiatric}. 
Traditional suicide ideation detection approaches rely on the knowledge of psychologists and self-reported questionnaires~\cite{4liu2020suicidal}. Patient Health Questionaire-9 (PHQ-9) and Columbia Suicide Severity Rating Scale (C-SSRS) are two examples of public forum questionnaires that can screen for suicide and identify depressive symptoms~\cite{5weber2017psychiatric}. These approaches are effective and quick, but they are subject to false negatives due to participant concealment. They are also difficult to carry out over a lengthy period or on a very large scale~\cite{5weber2017psychiatric}.

The task of identifying suicidality has attracted researchers in different fields to investigate linguistic and psychological signs and other factors that aid in diagnosing and identifying individuals with suicidal thoughts~\cite{4liu2020suicidal}.
Social media posts provide a valuable source of information about individuals' lives and their emotional and psychological states.
For various reasons, many individuals are unable to share their personal stories and express their emotions in real life and instead choose to write blogs about their feelings or suicide plans. Unfortunately, these suicide posts are often either overlooked or ignored. This information can help to perform screening of suicidality on a wide scale.

To detect suicidal individuals or who may have suicidal thoughts from their tweets or blogs is very important, because early detection of suicidal people could save many lives even though people who know that they are suffering from suicidal thoughts may not get the appropriate treatment for many reasons. Therefore, using a suicidal detection system could help many people and can have a significant impact on their treatments.

The studies reviewed in this paper have examined social media content to detect automatically suicidal ideation and behaviors. 
This article presents a detailed overview of current research efforts in social media platforms that use machine learning techniques to detect and identify suicidal ideation. Several specific tasks and datasets are introduced and summarized according to their practice. This article is intended for researchers who are interested in developing applications that leverage text classification methods or suicidal text classification. Also, to aid future study in the field and  investigate the feasibility of using social media to detect suicidal ideation.  In this research, the terms suicidal ideation, suicidal thoughts, and suicidality will be used interchangeably. The contributions of our survey are summarized as follows.

\begin{itemize}
\item To the best of our knowledge, this is the first comprehensive review of research into suicidal ideation detection using social media, including the datasets that have been constructed and the methods employed from a machine learning perspective.
\item We introduce and discuss classical and modern machine learning techniques on different social media platforms and identify the best performing algorithm in the context of the platform used in the study and how the dataset was collected and annotated.
\end{itemize}

\begin{figure}[ht]
\centering
\includegraphics[width=\linewidth]{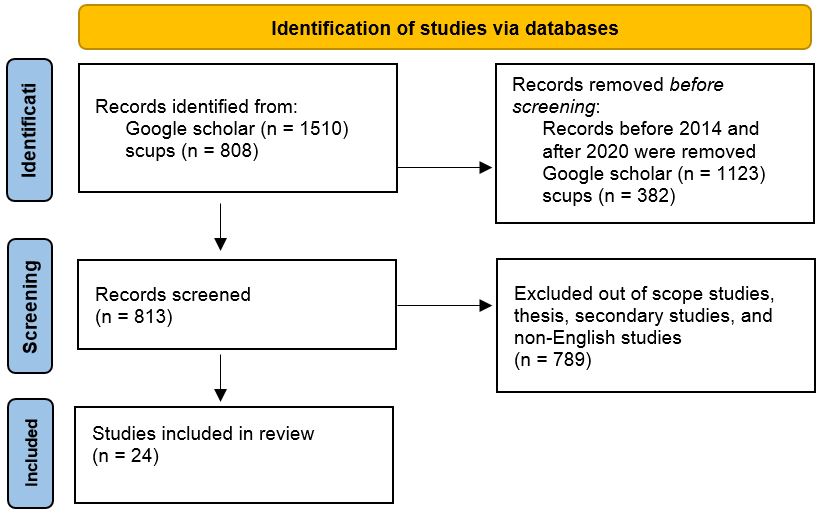}
\caption{Flow diagram for a systematic reviews which included searches of databases}
\label{fig:diagram}
\end{figure}

The literature search was performed through two databases for retrieving scientific works: Scopus and Google Scholar. These databases include most of the important papers in the area. The inclusion and exclusion criteria is shown in fig~\ref{fig:diagram} and can be summarized as follow. First, we included all papers from 2014 to 2020 that contain the following keywords in its title: (suicide OR suicidal OR suicidality OR suicide-related OR behavior OR ideation OR intent OR risk OR psychiatric stressors OR expressions OR detection OR detecting OR prediction) AND (deep OR machine OR learning OR algorithms OR classification OR feature selection OR social media OR Twitter OR Facebook OR Reddit OR Microblogs OR online communities). We then excluded out of scope studies, thesis, secondary studies (e.g., surveys, systematic literature reviews), and papers that had been written in a language other than English. 

The remainder of this article is organized as follows. Sections~\ref{Sec:Datasets} and ~\ref{Sec:Type of Data} detail the dataset collection procedures followed in the current research studies and annotation techniques. Section~\ref{Sec:methods} covers details of feature extraction and algorithms used in the classification process. Section~\ref{Sec:summary} provides a summary and discussion of the current research in the field. Section~\ref{Sec:conclusion} gives a conclusion of the survey paper. 

\section{Datasets}
\label{Sec:Datasets}
Users' posts and interactions on social media platforms provide a wealth of information for many researchers. Several sets of information, social media platforms, and data sources were investigated to identify suicide-related posts. This section gives an overview of current practice in the detection of suicidal thoughts. In particular, an overview of types of data (i.e., linguistic/semantic, psycholinguistic, metadata or interaction data), the language of the content (i.e., English, Chinese, and others), social media platforms (i.e., Twitter, Reddit) data collection procedure (including search keywords) and annotation scheme (i.e., number of classes) are given.

\subsection{Type of Data}
\label{Sec:Type of Data}

The studies surveyed in this paper examined several types of data categorized into linguistic data, psycholinguistic data, metadata, and interaction data~\cite{doi:10.1177/0261927X09351676}. Linguistic data was central to a series of NLP applications and includes,for example, authorship attribution and forensic linguistics, gender detection, and
personality type detection~\cite{danescu2011mark}. Many studies show that the linguistic and semantic features of social media users' posts could help indicate and clarify the mental state of the poster~\cite{10sawhney2020time}. Mapping words often obtain psycholinguistic features words into pre-defined psychological and affective categories. The Linguistic Inquiry Word Count (LIWC) is one of the most widely used psycholinguistic dictionaries in related NLP tasks\cite{doi:10.1177/0261927X09351676}. The LIWC consists of a large number of words along with different categories started by two effective classes (positive, negative emotion) and more than $80$ categories (e.g., anxiety, anger, sadness)~\cite{danescu2011mark,doi:10.1177/0261927X09351676}. The LIWC has been used in different domains such as social relations and mental health~\cite{danescu2011mark,6shah2020hybridized}.
Metadata features are pieces of information that describe digital data, which can be account metadata or post/message metadata. Account metadata are the data that describe the account, such as the owner's name, profile information, biography, and location. Post or message metadata are the data that describe posts, such as the author, location, likes, number of shares, date/time, links, and hashtags. Interaction data are associated with what users produce in their daily interactions and communication in the digital world~\cite{perez2018you}. Several interactive features were examined, which include user temporal posting patterns.

\subsection{Languages of Textual Data}

\begin{figure}[ht]
\centering
\includegraphics[width=\linewidth]{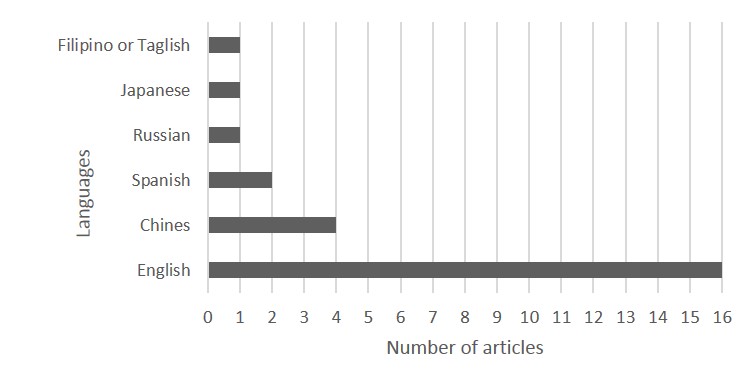}
\caption{Distribution of Articles over Languages.}
\label{fig:languages}
\end{figure}

Authors have examined the mental state of social media users in many languages.The majority of papers in the field are written in English \cite{8jain2019machine,10sawhney2020time,11o2015detecting,12valerianodetection,13burnap2015machine,14vioules2018detection,15moulahi2017dare,16ma2020dual,17huang2014detecting,18huang2019suicidal,23rabani2020detection,24ramirez2020detection,25chiroma2018text,27huang2015topic}. The Chinese language was the second most used language in the published studies~\cite{19narynov2019comparative,20du2018extracting,21tadesse2020detection,28fahey2020covariance}. Further studies were completed in Spanish~\cite{9ji2018supervised,26chiroma2018suiciderelated}, and Russian \cite{22rajesh2020suicidal}, and even Japanese \cite{24ramirez2020detection} and Filipino or Taglish \cite{1astoveza2018suicidal}.The distribution of articles over the platforms can be observed in Figure~\ref{fig:languages}. As Figure~\ref{fig:languages} shows, English-language articles predominate; of 24 articles, only seven used other languages.

\subsection{Platforms}
\begin{figure}[ht]
\centering
    \includegraphics[width=\linewidth]{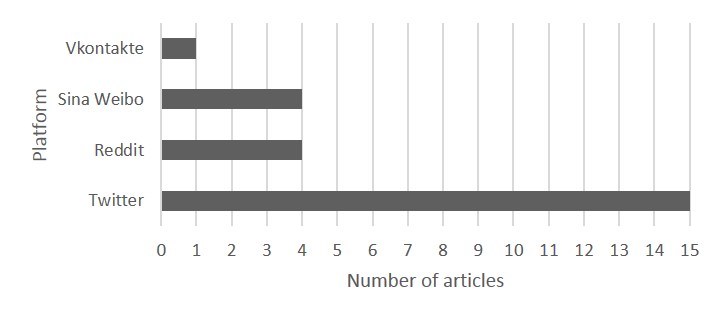}
    \caption{Distribution of Articles over Platforms}
  \label{fig:platforms}
\end{figure}

A wide range of social media platforms has been used for creating suicidality detection datasets, with most of the studies using Twitter~\cite{mishra-etal-2019-snap,10sawhney2020time,11o2015detecting}. Twitter is a free social media broadcast site, and any registered user can communicate with other users using 140 characters each time they post. Other social media platforms have been the subject of similar research, including Reddit~\cite{6shah2020hybridized,20du2018extracting}. Reddit is a community-driven platform for commenting, submitting,  and rating links and text posts~\cite{Redditsinger2014evolution}. The Chinese microblog Weibo has been studied~\cite{27huang2015topic,16ma2020dual,18huang2019suicidal,17huang2014detecting}, Weibo also has a limit of 140 characters in a post and has witnessed exponential growth, particularly in China~\cite{weibozhang2012motivations}. In Russia, a popular platform is Vkontakte in which users can create groups and invite users to join them, discuss different topics, and meet other users~\cite{19narynov2019comparative,Vkontaktesuleymanova2009tatar}. Figure~\ref{fig:platforms} shows the distribution of articles over these platforms and shows that, Twitter is the most used platform in studying suicidal posts.  

\subsection {Data Collection Procedure and Annotation schema}
Several datasets were developed for suicidality detection that vary in size, target (individual tweets or user histories), and data collection procedures. Sawhney et al. used the Twitter timeline dataset of user data~\cite{mishra-etal-2019-snap} to filter $32,558$ user profiles with a mean number of tweets history of $748$ tweets. A lexicon of $143$ suicidal phrases was used and then annotated by two clinical psychology students as Suicidal Intent (SI) Present or Suicidal Intent (SI) Absent~\cite{10sawhney2020time}. O'dea et al. gathered $1,820$ tweets to study suicide-related posts using English words or phrases consistent with the vernacular of suicidal ideation. Each tweet was then classified by three mental health researchers and two computer scientists~\cite{11o2015detecting}. Valeriano et al. collected $2,068$ Spanish tweets by translating a list of English keywords used to express a suicidal tendency to Spanish and then annotated the tweets with the help of bilingual assistants~\cite{12valerianodetection}. Burnap et al. collected four million tweets using suicidal keywords extracted from four well-known websites dedicated to suicide prevention and support. The dataset was then annotated using a crowd-sourcing online service and randomly sampled with $800$ suicidal tweets and $200$ undirect suicidal ideation tweets~\cite{13burnap2015machine}. Vioules et al. proposed an approach to detecting suicidal thoughts to identify sudden changes in users' online behavior by analyzing users' behavioral and textual features. They collected $5,446$ tweets using special key phrases obtained from a generated list of suicide risk factors and warning signs. Eight researchers and a mental health professional then
manually annotated tweets~\cite{14vioules2018detection}. 

Moulahi et al. exploited a list of key phrases generated from the American Psychological Association (APA) list of suicide risk factors and keywords from the American Association of Suicidology (AAS) list of warning signs. They only considered users' accounts that show in their online behavior serious suicide symptoms, collect $29,887$ tweets from $60$ users. To avoid over-fitting, they included $60$ normal accounts that used the same keywords~\cite{15moulahi2017dare}. Sawhney et al. extracted $4,314$ posts from four well-known Suicide web forums to create a suicidal language. Also, user posts with 'suicide' tag from social media sites such as Tumblr and Reddit were included to the collection. As a result, 300 posts were chosen from each suicide forum, and 2000 posts were chosen at random from Tumblr and Reddit. After manually annotating these posts and utilizing Term Frequency/Inverse Document Frequency (TF-IDF) to determine the most often occurring terms, a list of 108 words/phrases associated with Suicidal Intent was created. To validate the model's performance in terms of various elements, three datasets were constructed using different strategies: ($2726$ suicidal, $9160$ non-suicidal) using words/phrases. The second dataset followed the same method and users whose tweets or posts were classified as suicidal but didn't include any hashtags associated with suicidal ideation, and the last dataset used both datasets with no overlap. To assess the effectiveness of the proposed methodology, three clinical psychology students annotated each of the three datasets, which included suicidal and nonsuicidal tweets.~\cite{Lnew8858989}. Astoveza et al. gathered a dataset using keywords of potential warning signs and hints from psychological associations and online organizations and keywords used in similar studies. The chosen keywords were translated to the Filipino language to gather $3,055$ English and $2,119$ Filipino or Taglish tweets and annotated by trained psychologists and a resident guidance counselor~\cite{1astoveza2018suicidal}.

Shah et al.used Reddit to gather $7,098$ English posts. The dataset consisted of $3,549$ user-posts containing suicidal ideation taken from a sub-Reddit called SuicideWatch and labeled "1". A further $3,549$ pieces of data of different popular Reddit posts that do not contain suicidal ideation and labeled "0"are also included~\cite{6shah2020hybridized}. The dataset consists of $594$ suicidal ideation tweets out of $10,288$ tweets using a keyword filtering technique including suicidal words and phrases such as, e.g., suicide, die, and end my life. The text is then manually annotated to Suicide, Nonsuicide text~\cite{9ji2018supervised}. Questionnaires are also considered textual data sources. Jain et al. used two datasets, one from questionnaires and the second is from Reddit and Twitter and a labeled dataset from Kaggle~\cite{8jain2019machine}. 

In2012,a Chinese college student, Zoufan, hung herself after leaving a suicide note on Weibo, the largest open social media platform in China. People still paid attention and left messages below her last blog, with some of the messages reveal suicidal thoughts.Y. Huang et al. created a dataset by sorting through $65,352$ messages below Zoufan's last blog entry. Three experts who specializing in psychology and suicidal behavior labeled $8,548$ blogs as suicide and $10,000$ as non-suicide blogs~\cite{18huang2019suicidal}. A further, another dataset used Zoufan's blog and crawled $5,000$ Chinese posts from the Weibo website to be used in Dual attention mechanism (DAM) to improve the performance of social media based suicide risk detection~\cite{16ma2020dual}. Huang et al. identified $53$ users who posted suicidal content on Weibo before their deaths and collected more than $30,000$ posts, in addition to another, they also collected $600,000$ posts collected from $1,000$ thousand random non-suicidal users. The researchers curated all suicidal users' posts and obtained $614$ suicidal posts. They then randomly sampled $6,140$ posts from the set of non-suicidal users for a total of $6,754$ posts. After filtering some blank posts, they obtained $6,704$ posts~\cite{17huang2014detecting}. 

Researchers in another study acquired messages from VKontakte, Europe's second-largest social network after Facebook. They gathered $35,000$ Russian messages from individuals diagnosed with depression (i.e., chronic, severe, and persistent), and $50,000$ postings with unfavorable sentiments on other topics were obtained to generate a balanced dataset~\cite{19narynov2019comparative}. 

\begin{figure}[ht]
\centering
    \includegraphics[width=14cm,height=5cm]{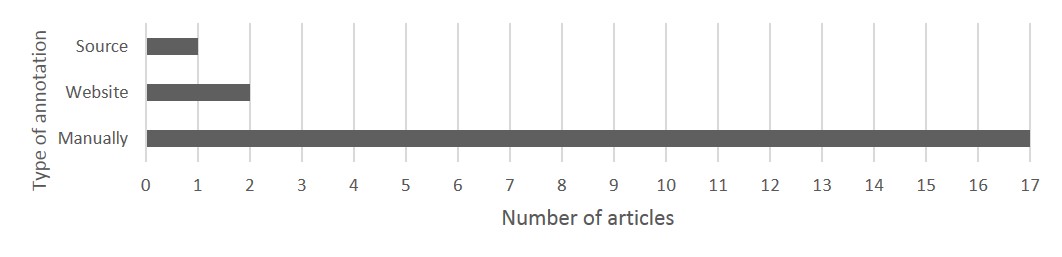}
    \caption{Annotation Scheme used by Articles}
  \label{fig:annotation}
\end{figure}

Figure ~\ref{fig:annotation} shows the frequency usage of each annotation type and shows that the data is annotated manually in most studies. They could be students of clinical psychology or mental health researchers. Two studies used a website like Kaggle to annotate their data. In one study, the data were annotated based on the source~\cite{6shah2020hybridized}, further explained in the next section.   

\subsection{Annotation Classes}
~\label{Sec:Annotation}
Data annotation is an important aspect of data preparation because supervised machine learning models spot patterns in annotated data. Each example in the dataset should be labeled with one of the pre-defined categories to train machine learning models to distinguish between potential suicide ideation and nonsuicide ideation posts. 

\begin{figure}[ht]
\centering
    \includegraphics[width=14cm,height=6.3cm]{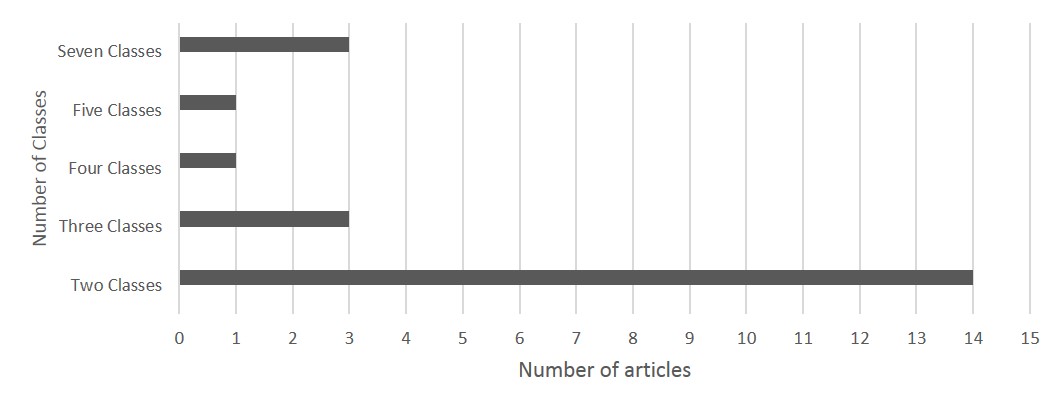}
    \caption{Number of class used in Classification in each article}
  \label{fig:Classes2}
\end{figure}

The problem of suicidal ideation detection is often formulated in binary and multiclass classification. Binary suicidal classification determines whether a given post contains suicidal thoughts or a user is at risk of suicide. Figure~\ref{fig:Classes2} shows that in most studies data is labeled as a binary classification of 0 (non-suicidal), or 1(suicidal)~\cite{6shah2020hybridized,10sawhney2020time,18huang2019suicidal,20du2018extracting,21tadesse2020detection,22rajesh2020suicidal,Lnew8858989,1astoveza2018suicidal,12valerianodetection}. Shah et al. labeled data according to the source assigning the label "1" to the data that are from Suicide-Watch and "0" to the data from other sub-Reddit forums~\cite{6shah2020hybridized}. Jain et al. labeled the second dataset (Reddit and Twitter) using two classes, risky and non-risky and annotated the first dataset (questionnaire) using five levels of depression severity~\cite{8jain2019machine}.
Multiclass classification tasks are formulated based on the assumption that each sample is assigned to several pre-defined classes. For example, O'Dea et al. classified posts into three levels: "Strongly concerning," "Possibly concerning," and "Safe to ignore." Annotators were instructed to choose only one of the three levels and, if in doubt, to choose "Safe to ignore"~\cite{11o2015detecting}. Four levels of classification were used by Vioules et al. ranging from normal to suicidal~\cite{14vioules2018detection}. Burnap et al. classified Twitter text into seven classes, including suicidal intent, or other suicide-related topics such as suicide campaigning, support or prevention of suicidality, reporting of suicide, flippant reference to suicide, or none of the above~\cite{13burnap2015machine,25chiroma2018text,26chiroma2018suiciderelated}.

\section{Methodology}
~\label{Sec:methods} 
\begin{figure}[ht]
\centering
   \includegraphics[width=11cm,height=4cm]{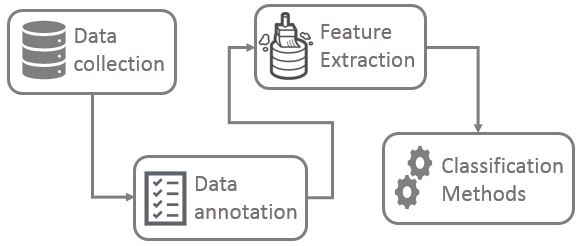}
    \caption{Architecture of Suicide Detection Methodology}
  \label{fig:steps}
\end{figure}

The classification of suicidal-related posts or blogs aims to determine whether the user has a suicidal tendency or not. Machine learning methods and other techniques have also been applied to solve this problem. The classification method often requires employing feature extraction/ text representation technique before employing machine and deep learning models.  
Figure~\ref{fig:steps} shows a general procedure used by most studies discussed in this article.
Step one was data collection and involved constructing a dataset using one or more social media platforms. The second step, annotation, involved labeling datasets using different techniques, as discussed in section~\ref{Sec:Annotation}. The third step, is feature extraction, is applied before employing machine and deep learning models.

\subsection{Feature extraction}
~\label{Sec:feature}
Many techniques have been used to extract features from social media posts to identify whether they reflect suicidal thoughts or not.

TF-IDF matrixes were used for textual features to reflect the importance of words to distinguish between suicidal and non-suicidal posts~\cite{8jain2019machine,11o2015detecting,19narynov2019comparative,12valerianodetection,18huang2019suicidal,23rabani2020detection,Lnew8858989}. N-gram features were also utilized to find the probability of $n$ words in a given document:in this case  utilized to process blog content and identify terms in the blog corpus.~\cite{18huang2019suicidal,Lnew8858989,1astoveza2018suicidal}. N-grams are known as a base feature for sentiment analysis of tweets. Due to the character limitation in Twitter posts, it leads to choosing short N-grams~\cite{14vioules2018detection}.

Some studies use textual features in addition to psycholinguistic features obtained from (LIWC)\cite{doi:10.1177/0261927X09351676}. LIWC is also used to count the frequency of a specific word, and LIWC has categories to identify syntactical elements (e.g.,nouns, pronouns, verbs, and adverbs)~\cite{24ramirez2020detection,18huang2019suicidal,Lnew8858989}.

Computational features and linguistic features (TF-IDF, N-gram and 30 best features for LIWC) were used by Shah et al. to propose a hybrid method~\cite{6shah2020hybridized}. Several features were extracted, including statistical, linguistic, syntactic, topic features, and word embedding. These features were used to detect online suicidal users through their online content~\cite{9ji2018supervised}, including the language of the tweet and the emotional historic spectrum feature in a time-aware manner~\cite{10sawhney2020time}. Important data that can give information about users on social networks include users’ behavior (daily activities, social network size, etc.)~\cite{14vioules2018detection}. Profile, text content such as messages, publications, and comments are used to extract features of a general-purpose classification~\cite{19narynov2019comparative}. A combination of textual features such as BoW or N-grams and word embedding with social network and psychological features include lexical, behavioral, and sentiment analysis. These and other features can be mapped to social media context using certain signs, symptoms, and image-based features~\cite{24ramirez2020detection}.  

\subsection{Classification Methods}
Many studies have utilized machine classification techniques to study and analyze the content users generate on social media. First, researchers have focused on three strategies to tackle the problem of suicide detection. Researchers formulated the problem as a time-series problem to detect changes in users' behavior. Second, the task is see as a text classification (supervised) problem to identify linguistic connotations associated with suicide.Third, the problem is approached as unsupervised (clustering)  to group examples of user posts into different groups based on their features.

Several supervised algorithms were examined in the literature, including 
Support Vector Mchine (SVM)~\cite{6shah2020hybridized,8jain2019machine,24ramirez2020detection}, NB~\cite{6shah2020hybridized,26chiroma2018suiciderelated,23rabani2020detection}, K-nearest neighbor algorithm (KNN)~\cite{6shah2020hybridized,14vioules2018detection},
Logistic Regression 
(LR)~\cite{8jain2019machine,11o2015detecting,24ramirez2020detection}, Decision Tree (DT) classifier~\cite{8jain2019machine,27huang2015topic}, and Extreme Gradient Boost (XGBoost) algorithm~\cite{8jain2019machine,22rajesh2020suicidal,9ji2018supervised}. Time-aware Long Short-Term MemoryLSTM (T-LSTM) was used to propose Suicidality Assessment Time-Aware Temporal Network (STATENet)\cite{10sawhney2020time}. Convolutional Neural networks (CNN) and Recurrent Neural Networks (RNN) were also used to classify suicidal posts.

\subsubsection{Temporal Behavior Problem}
 The Multi-Layer Perceptron classifier was also used with $1,500$ best features out of $5,000$ features. The classifier was able to classify $90.2\%$ of the non-risky tweets and only misclassified $9.0\%$. However, only $65.1\%$ of the risky tweets were classified correctly   ~\cite{1astoveza2018suicidal}.
A study in the Japanese language used ordinary least squares (OLS) regression model to study the relationship between suicide cases and the suicidal keyword “kietai” (“I want to disappear”). The researchers also studied the linguistic context changes at different hours of the day for the suicidal keyword. They found a clear pattern with the use of suicidal keywords peaking from 1 am to 5 am. This trend showed a positive correlation among suicide deaths for people aged 15 to 44 years but negative among adults over 45 years old. Nighttime tweets showed a significant relationship between self-disgust words and words that indicate direct suicidal ideation~\cite{28fahey2020covariance}.

A probabilistic framework based on Conditional Random Fields (CRF) was used by Moulahi et al. to track suicidal ideation. They studied mental states as a sequence of events, considering the context and users’ online activities that may lead to suicide. They evaluated their approach by comparing it with other machine learning methods: SVM, NB, J48, RF, and multilayer perceptron. Different CRF configurations were run, and no sequences of observations were considered to compare their approach. The researchers noted that both CRF configurations outperform in terms of all the testing criteria average Precision, recall, and F1-score measures. Their approach had the best performance Precision of $81.6\%$, recall $75.2\%$, and F1-score $71.1$~\cite{15moulahi2017dare}. The Firefly algorithm is a metaheuristic-based approach that seeks to increase classifier accuracy while attempting to reduce the amount of features in order to reduce computational cost, complexity, and redundancy. Sawhney et al. used the Binary Firefly Algorithm (BFA) which is a discrete-space modification of the firefly algorithm used for feature selection. They used firefly algorithm as a wrapper over the four classifiers (Random Forest (RF), SVM, LR, and XGBoost). RF and BFA combined gave the highest performance with $89.2\%$ Precision, $87.4\%$ recall, and $88.3\%$ F1-score\cite{Lnew8858989}

Vioules et al. detect the change in the data streams by passing textual and behavior features to a martingale framework. They needed two datasets sufficiently large annotated set and another smaller set of selected Twitter users to study their history. They found that the two-step classification performed well in the test set. They reached $82.9\%$ precision, $81.7\%$ for recall, and F1-score\cite{14vioules2018detection}. 

A DAM finds the correlation between text and image from the same post and better detects the user’s implicit suicide risk. They have then compared their model with other five models: NB, SVM with TF-IDF features, Long Short-Term Memory (LSTM), CNN, and Species Distribution Models (SDM) deep learning model based on layered attention and suicide-oriented word embeddings. Experiments showed that the DAM performed better than most suicide risk detection models and obtained competitive results on the proposed dataset. The model performed better when people’s posts contained images~\cite{16ma2020dual}.

\subsubsection{Text Classification Problems}
Classification algorithms such as SVM and LR were examined to identify a tweet with a tendency to suicide from a non-suicidal tweet~\cite{12valerianodetection}. 
Narynov et al. used supervised (Gradient Boosting, RF) and unsupervised algorithms (K-means) and tested them with TF-IDF and Word2Vec. They found that RF with TF-IDF had the best performance with $96\%$ accuracy~\cite{19narynov2019comparative}. 

Six supervised learning classifiers were used: SVM, RF, gradient boost decision tree (GBDT) for classification, XGBoost, and feed-forward neural network with several sets of features (statistics, POS counts, LIWC features, TF-IDF vectors, and topic probability features) and found that combining more features increases the performance of all methods. RF gained better performance than most models except for the metric of Precision, in which the neural network model achieves slightly better results~\cite{9ji2018supervised}. Different classifiers were also examined by X. Huang et al., including SVM, NB, LR, J48 classifier, RF, and Sequential minimal optimization (SMO) with three N-gram features. SVM classifier achieves the best performance in comparison with other classifiers, with an F1-score of $68.3\%$, a Precision of $78.9\%$, a recall of $60.3\%$, and accuracy over $94.0\%$ ~\cite{17huang2014detecting}.

Different machine learning algorithms and ensemble approaches have been used, such as NB, decision trees, multinomial NB, LR, RF, resulting in $98.5\%$  accuracy, $98.7\%$ Precision, and $98.2\%$  recall yielded using RF that gave the best performance~\cite{23rabani2020detection}.

Two machine learning algorithms were used (SVM, DT) by Y. Huang et al. to build a classification model with three features sets (automated machine learning dictionary, Chinese suicide dictionary, and Simplified Chinese Micro-Blog Word Count (SCMBWC)). Each feature set was used with two machine learning algorithms separately to generate six detection results. Those were input to an LR. It has been found that SVM with feature set extracted using automated machine learning dictionary from real blog data-driven by N-gram had the best performance~\cite{18huang2019suicidal}. 
 
Tadesse et al. combined two models, LSTM and CNN, to explore the potential of each algorithm separated and their combined model applied in classifying the sentences with suicidal and non-suicidal content. The proposed algorithm was compared with CNN, LSTM separated and other machine learning classifiers such as SVM, NB, RF, and XGBoost. They found the proposed model improved the accuracy with $93.8\%$, F1-score $93.4\%$, recall $94.1\%$ and Precision $93.2\%$ \cite{21tadesse2020detection}.

SVM and NB were incorporated as an ensemble approach known as Rotation Forest (RF). They tested the RF approach with three classifiers: DT, SVM, and NB. They reached $69.0\%$ for F1-score, the Precision performance of $64.4\%$, and recall of $74.4\%$\cite{13burnap2015machine}. Interestingly, another study~\cite{25chiroma2018text} used four machine classifiers (DT, NB, RF, and SVM) on the same dataset~\cite{13burnap2015machine} and DT had the best performance with an F1-score of $87.9\%$ and $79.0\%$ accuracy for a multiclass dataset. A third study completed by Chiroma et al. made the same dataset~\cite{13burnap2015machine} with the same pre-processing technique. The prism algorithm was first introduced in 1987 by Cendrowska \cite{prismCENDROWSKA1987349}. It can select attributes based on their importance to a specific class \cite{26chiroma2018suiciderelated}. They compared the performance of the Prism algorithm against the popular machine learning algorithms (SVM, DT, NB, and RF). They found that the Prism algorithm had $84\%$ Precision, recall, and F1-score, which is the best performance compared to all other classifiers in all measures\cite{26chiroma2018suiciderelated}.

Sentiment dictionaries were adopted into Latent Dirichlet Allocation (LDA) by X. Huang et al. and evaluated against traditional LDA on a different number of topics (100- 1000). Also, they trained and tested different classifiers SVM, J48 classifier, LR, random tree, RF, and decision table. They found the best performing algorithm was the J48 classifier with an accuracy of 94.3, Precision $80.2\%$, recall $48.3\%$, and F1-score $60.3\%$~\cite{27huang2015topic}.

Vader sentiment analysis was used by Rajesh Kumar et al. to give a score for each word with different classifiers such as NB, RF, XGBoost, and logistic regression. Vader sentiment analysis helped separate the sentence to distinguish the sentences into positive, or neutral. They achieved $99.6\%$ accuracy using the RF method~\cite{22rajesh2020suicidal}. CNN to select suicide-related tweets and RNN to extract stressors were used by Du et al. to build an automatic psychiatric stressors binary classifier. They compare their proposed model with other machine/deep learning approaches SVM, ET, RF, LR, Bi-LSTM. CNN had the highest recall: $90\%$ and F1-score:$83\%$~\cite{20du2018extracting}.The studies are summarized in Table 1.

\newpage
\onecolumn
\small
\begin{center}

\begin{longtable}{ c c c c c c c} 
\caption{Results of Each Study Included in This Review. * indicates best performing algorithm. Acc,P,R, and F1 are abbreviation for Accuracy, Precision, Recall, and F1-score, respectively.SNPSY: social networks and psychological features.} \label{tab:long} \\

\hline \multicolumn{1}{c}{\textbf{Ref/Year.}} 
& \multicolumn{1}{c}{\textbf{Source}} & 
\multicolumn{1}{c}{\textbf{language}} &
\multicolumn{1}{c}{\textbf{N(Posts)}}&
\multicolumn{1}{c}{\textbf{Features}}&
\multicolumn{1}{c}{\textbf{Algorithms}}&
\multicolumn{1}{c}{\textbf{Performance}}
\\ \hline 
\endfirsthead

\multicolumn{7}{c}%
{{\bfseries \tablename\ \thetable{} -- continued from previous page}} \\
\hline \multicolumn{1}{c}{\textbf{Ref/Year.}} &
\multicolumn{1}{c}{\textbf{Source}} & 
\multicolumn{1}{c}{\textbf{Language}} &
\multicolumn{1}{c}{\textbf{N(Posts)}}&
\multicolumn{1}{c}{\textbf{Features}}&
\multicolumn{1}{c}{\textbf{Algorithms}}&
\multicolumn{1}{c}{\textbf{Performance}}
\\ \hline 
\endhead

\hline \multicolumn{7}{r}{{Continued on next page}} \\ \hline
\endfoot

\hline \hline
\endlastfoot

\cite{6shah2020hybridized}/2020	& 	Reddit	&	English	&\makecell{	7098\\ post} &\makecell{		Unigram,\\Bigram,\\Trigram,\\ TF-IDF,\\ LIWC}&	\makecell{NB*,\\ SVM,\\ KNN,\\ RF}	&	
\makecell{
\newline Acc:73.6\%\\
\newline P:70.5\%\\
\newline R:89.7\%\\	
\newline F1:76.7\%
}
\\\hline 	

\cite{8jain2019machine}/2019 &	\makecell{Twitter\\ Reddit\\ qustinayr}	&	English	&	-		&	TF-IDF	&\makecell{	LR*,\\ DT,\\ XGBoost,\\ SVM}& 86.5 \\ \hline

\cite{10sawhney2020time}/2020 &	Twitter	&	English	& \makecell{	34,306\\ tweets}	& \makecell{	LIWC,\\ N-grams ,\\POS }	&\makecell{	RF,\\ LSTM,\\ SDM, \\CNN,\\ STATENet*}	&
\makecell{
\newline Acc:85.1\%\\	
\newline R:81.0\%	\\
\newline F1:79.9\%	
}
\\ \hline

\cite{19narynov2019comparative}/2019 &	VKontakte	&	Russian	&	\makecell{85,000\\ posts} 	&		\makecell{TF-IDF,\\ Word2Vec}	&\makecell{	GB*, RF,\\ k-means}	
& 
\makecell{
\newline P:96.0\%	\\
\newline R:95.0\%	\\
\newline F1:95.0\% 
}
\\ \hline

\cite{24ramirez2020detection}/2020 &	Twitter	&	Spanish	&	\makecell{1200\\ users}	& \makecell{LIWC, \\BoW, \\N-gram,\\ SNPSY,\\images} &\makecell{	RF, LR,\\MLP, \\SVM* ,\\ CNN}	&
\makecell{
\newline Acc:86.0\%\\
\newline P:91.0\%	\\
\newline R:81.0\%	\\
\newline F1:86.0\%
}
\\ \hline

\cite{23rabani2020detection}/2020 &	Twitter	&	English	&	\makecell{4266\\ tweets}		&	\makecell{TF-IDF, \\ BoW}	&\makecell{	NB, DT,\\ SVM,\\ RF*, LR,\\ and others}
& 
\makecell{
 Acc:98.5\%	\\
 P:98.7\%\\
 R:98.2\%
}
\\ \hline

\cite{12valerianodetection}/2020 &	Twitter	&	Spanish	&	\makecell{2068\\ tweets}	&		\makecell{TF-IDF,\\ Word2Vec.}	& \makecell{	SVM,\\ LR*}	&
\makecell{
 Acc:79.0\%\\
 P:79.0\%\\
 R:79.0\%\\
 F1:79.0\%
}
\\ \hline

\cite{27huang2015topic}/2015 &	Weibo	&	Chinese	& \makecell{7978\\ blogs}		& \makecell{	Word2Vec,\\ POS, LDA,\\ meta\\ features, \\N-gram	}	&\makecell{	SVM, J48*,\\ LR, RT,\\ RF, DT}	&	
\makecell{
 Acc:94.3\%	\\
 P:80.2\%	\\
 R:48.3\%	\\
 F1:60.3\%	
}
\\ \hline

\cite{25chiroma2018text}/2018	&	Twitter	&	English	&	\makecell{1000\\ tweets}	& \makecell{POS,\\ BOW,\\TF-IDF}	&	\makecell{DT, NB, \\RF, SVM}	&	
\makecell{
 P:86.4\%\\
 R:89\%	\\
 F1:87.9\%	
}
\\ \hline

\cite{26chiroma2018suiciderelated}/2018	& 	Twitter	&	English	&	\makecell{1000\\ tweets}	& \makecell{TF-IDF,\\ N-gram,\\ POS,\\ BOW,\\ LIWC}	&	\makecell{Prism\\ algorithm*,\\ DT, NB,\\ RF, SVM}	& 
\makecell{
 P:84.0\%\\	
 R:84.0\%\\	
 F1:84.0\%
}
\\ \hline	

\cite{22rajesh2020suicidal}/2020 	&	Twitter	&	English	&	\makecell{54720\\ tweets}		& \makecell{ statistical,\\\ BOW,\\Word \\frequency}	&	\makecell{NB,\\ RF*,\\ LR\\ XGBoost  	}&
\makecell{
 P:99.6\%\\
 R:99.1\%\\
 F1:99.8\%
}
\\ \hline

\cite{16ma2020dual}/2020 	&	Weibo	&	Chinese	&	\makecell{5,000 \\users}		&	TF-IDF	&	\makecell{SDM, CNN,\\ LSTM, NB,\\ SVM, DAM*	}&
\makecell{
\newline Acc:91.8\%\\	
\newline F1:91.5\%\\
}
\\ \hline

\cite{11o2015detecting}/2015 	&	Twitter	&	English	&	\makecell{1820\\ tweets}		&	\makecell{ freq,\\ TF-IDF,\\ filter}	&SVM* ,LR	&	
\makecell{
 Acc:76.0\% \\
 P:80.0\% \\
 R:53.0\%\\
 F1:64.0\%
}
\\\hline	
\cite{18huang2019suicidal}/2019 &	Weibo	&	Chinese	&	\makecell{18548\\ blogs}	&	\makecell{ 	N-gram,\\TF-IDF,\\ LIWC	}&	SVM*, DT	&
\makecell{
 P:89.0\%	\\
 R:88.0\%\\
 F1:88.0\%
}
\\\hline

\cite{9ji2018supervised}/2018 & \makecell{ Reddit\\ Twitter}	&	English	&\makecell{	10882 \\tweets}	&	\makecell{Statistical, \\POS,\\ LIWC,\\ Word2Vec, \\ LDA}	&\makecell{	SVM, RF*,\\ GBDT, \\LSTM\\	and others}&	
\makecell{
 Acc: 96.4\% \\
 P: 96.4\% \\
 R: 99.2\% \\
 F1: 96.5\%	}
\\ \hline

\cite{21tadesse2020detection}/2019 & 	Reddit	&	English	&	\makecell{7201\\ posts}	& \makecell{ 	TF-IDF,\\ BoW,\\ Statistical,\\ Word2Vec}	&	\makecell{SVM, NB, \\RF, XGBoost, \\LSTM, CNN,\\ LSTM-CNN*}	& 
\makecell{
 Acc:93.8\%	\\
 P:93.2\%\\
R:94.1\%\\	
F1:93.4\%}	
\\ \hline

\cite{14vioules2018detection}/ 2018	&	Twitter	&	English	&	\makecell{5,446\\ tweets} &	\makecell{ N-grams,\\ symptoms,\\ pronouns, \\swear}	& \makecell{NB, SMO ,\\ J48, LR, RF, \\and others }
& 
\makecell{ P:83.0\%\\
R:82.0\% \\
 F1:82.0\%}
	\\ \hline

\cite{20du2018extracting}/2017 &	Twitter	&	English	&\makecell{	6,263\\ tweets}	&	\makecell{ GloVe\\ Twitter\\ embedding}	&	\makecell{CNN*, SVM,\\ ET, RF,LR,\\ Bi-LSTM}	&
\makecell{
 P:78.0\%\\	
 R:88.0\%\\	
 F1:83.0\%}
\\ \hline	
\cite{17huang2014detecting}/2014 &	Weibo	&	Chinese	&\makecell{	614\\ posts}&	\makecell{ Unigram,\\ Bigram,\\ Trigram}	&\makecell{	SVM*, NB,\\ LR, J48,\\ RF, SMO}	&
\makecell{
 Acc:94.0\%\\
 P:78.9\%\\
 R:60.3\%\\
 F1:68.3\%}
\\ \hline	

\cite{15moulahi2017dare}/2017 &	Twitter	&	English	&	\makecell{29887\\ tweets}	&	\makecell{POS ,\\ sentiment \\(Psychological\\ and\\
emotional \\lexicon)\\, contextual}	&\makecell{	SVM, NB, \\J48, RF,\\  DARE* \\and others.}	&
\makecell{ P:81.6\%\\
 R:75.2\%\\
 F1:71.1\%}
\\ \hline	
\cite{13burnap2015machine}/2015	 &	Twitter	&	English	&	\makecell{1000\\ tweets}	&\makecell{ TF-IDF,\\ N-gram,\\ POS, LIWC}	&\makecell{	NB, SVM,\\ J48, RF,\\ NB+SVM*}&
\makecell{ P:64.4\%\\
 R:74.4\%\\
 F1:69\%}
\\ \hline	
\cite{28fahey2020covariance}/2020	&	Twitter	&	Japanese	&	\makecell{2,889,190\\tweets} &		-	&	\makecell{OLS\\ regression}	&	
\\ \hline

\cite{Lnew8858989}/2019	&	Twitter	&	English	& \makecell{36548\\ tweets} &	 \makecell{Unigrams,\\ Bigrams,\\ LIWC,\\ TF-IDF, \\POS, LDA \\unigrams, \\Sentiment\\ Ratio,\\ Emoji\\ Sentiment} &\makecell{	RF, SVM,\\ LR, RNN,\\ LSTM, \\RF + BFA* \\and others}
	&	

 \makecell{P:89.2\%\\
 R:87.4\%\\
 F1:88.3\%}
\\ \hline	

\cite{1astoveza2018suicidal}/2018 &  	Twitter	& \makecell{English \\Filipino} & \makecell{5,174\\ tweets} 
&	\makecell{Unigrams,\\ Bigrams}	&	MLP &
 Acc:89.2\%
\\ \hline	

\end{longtable}
\end{center}
\newpage
\section{Discussion}
~\label{Sec:summary}
Detecting suicidal people using new technologies is an important and very active research area. Many studies have been developed to detect suicidal ideation using different machine learning techniques automatically. Users' posts and their interaction on different social media platforms is a novel area of inquiry. This review paper discusses different studies that use machine learning techniques on social media platforms to detect and identify suicidal ideation. Both supervised and unsupervised machine learning algorithms were used on different social media platforms such as Twitter, Reddit, and other microblogs, adopting different languages such as English, Chinese, Spanish, Japanese, and Russian, as shown in Table 1. Several datasets have been developed using different procedures for suicidal ideation detection purposes. The most commonly used procedures are keyword and suicidal phrases extracted from suicide dictionaries or translated from other languages, obtained from websites or lists of suicidal supports.
A subset of studies investigated metadata or interaction data, but most studies used linguistic data. Metadata can show how and when a person is active, indicating a person’s psychological state. Linguistic and sentiment analysis of users' posts also showed a good understanding of users' emotional and mental health.

\begin{figure}[ht]
\centering
   \includegraphics[width=\linewidth,height=8cm]{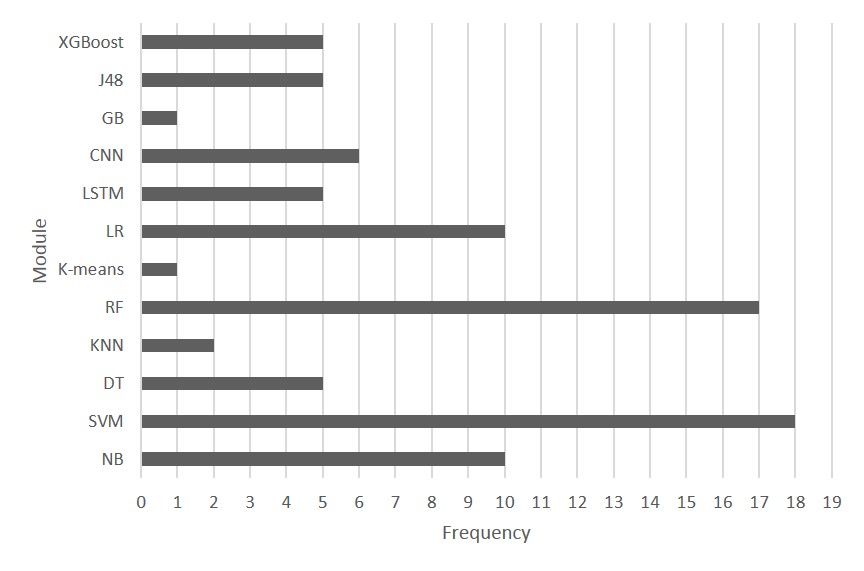}
    \caption{Distribution of Most used Module}
  \label{fig:Module}
\end{figure}

Most studies used and compared their work using popular machine learning algorithms such as LR, DT, SVM, RF, and NB. In other studies, deep learning algorithms like CNN and LSTMs were used. Figure~\ref{fig:Module} shows the frequency of module usage. SVM and RF are the most used models, and followed by LR and NB. The classification was most commonly observed in this review, with a small number of studies using time-frame, and other studies using both. Most studies classified posts, although some classified users. There are varying numbers of classes or labels for both classification types to determine the level of concern. Most studies used only two classes (suicidal and non-suicidal), although some used additional classes for uncertainty, and other studies used three to five levels. Different sets of features were used, including statistical, syntactic, linguistic, and topic features. Most researchers use different textual features such as TF-IDF, N-gram, and LIWC. Meta features were also used, like posting time and social relationships. Methods with automatic feature learning increased the performance of suicidal ideation detection. Table 1 provides an overview of all studies mentioned in this article.

\section{Conclusion}
~\label{Sec:conclusion}
Using social media platforms to express experiences and feelings has created new opportunities to analyze and detect suicidal ideation and other mental disorders. The early detection of suicidal ideation on social media networks will reduce suicide, provide an automatic and wide-ranging screening for suicidal tendencies and prevent the spread of suicidal content in social media. This survey investigates existing methods that use social media to detect suicidal ideation using machine learning methods. A significant amount of research has confirmed the effectiveness and feasibility of using social media such as Twitter, Reddit, and Weibo for suicidal ideation detection. Most studies have focused on suicidal ideation detection techniques for widely used languages such as English, but less attention has been paid to Arabic. Thus, with the growing number of social media users in the Arab region, research is needed for Arabic suicidal ideation detection. 

\newpage

\bibliography{sample}

\end{document}